\documentclass[sigconf,nonacm]{acmart}

\usepackage[linesnumbered,ruled,vlined]{algorithm2e}

\AtBeginDocument{%
  }

\acmConference[MM '25]{Proceedings of the 33rd ACM International Conference on Multimedia}{October 27-October 31, 2025}{Dublin, Ireland}

\acmBooktitle{MM '25: ACM International Conference on Multimedia,
October 27-October 31, 2025, Dublin, Ireland}

\usepackage{tcolorbox}
\usepackage{url}
\usepackage{xcolor}
\usepackage{colortbl}
\definecolor{mygray}{gray}{.9}
\newtcolorbox[list inside=prompt,auto counter,number within=section]{prompt}[1][]{
    colbacktitle=black!60,
    coltitle=white,
    fontupper=\footnotesize,
    boxsep=5pt,
    left=0pt,
    right=0pt,
    top=0pt,
    bottom=0pt,
    boxrule=1pt,
    title={#1},
    #1, 
}

\begin{document}

\title{M-MRE: Extending the Mutual Reinforcement Effect to Multimodal Information Extraction}

\author{Chengguang Gan}
\affiliation{%
  \institution{Yokohama National University}
  \city{Yokohama}
  \country{Japan}
}
\email{ganchengguan@yahoo.co.jp}

\author{Zhixi Cai}
\affiliation{%
  \institution{Monash University}
  \city{Melbourne}
  \country{Australia}
}
\email{zhixi.cai@monash.edu}

\author{Yanbin Wei}
\affiliation{%
  \institution{Southern University of Science and Technology}
  \city{Shenzhen}
  \country{China}
}
\affiliation{%
  \institution{Hong Kong University of Science and Technology}
  \city{Hong Kong}
  \country{China}
}
\email{yanbin.ust@gmail.com}

\author{Yunhao Liang}
\affiliation{%
  \institution{University of Chinese Academy of Sciences}
  \city{Beijing}
  \country{China}
}
\email{liangyunhao22@mails.ucas.ac.cn}

\author{Shiwen Ni}
\affiliation{%
  \institution{Shenzhen Institute of Advanced Technology, Chinese Academy of Sciences}
  \city{Shenzhen}
  \country{China}
}
\email{sw.ni@siat.ac.cn}

\author{Tatsunori Mori}
\affiliation{%
  \institution{Yokohama National University}
  \city{Yokohama}
  \country{Japan}
}
\email{tmori@ynu.ac.jp}


\newcommand{\czx}[1]{{\textcolor{orange}{#1}}}

\renewcommand\footnotetextcopyrightpermission[1]{}
\settopmatter{printacmref=false} 

\begin{abstract}

Mutual Reinforcement Effect (MRE) is an emerging subfield at the intersection of information extraction and model interpretability. MRE aims to leverage the mutual understanding between tasks of different granularities, enhancing the performance of both coarse-grained and fine-grained tasks through joint modeling. While MRE has been explored and validated in the textual domain, its applicability to visual and multimodal domains remains unexplored. In this work, we extend MRE to the multimodal information extraction domain for the first time. Specifically, we introduce a new task: Multimodal Mutual Reinforcement Effect (M-MRE), and construct a corresponding dataset to support this task. To address the challenges posed by M-MRE, we further propose a Prompt Format Adapter (PFA) that is fully compatible with various Large Vision-Language Models (LVLMs). Experimental results demonstrate that MRE can also be observed in the M-MRE task, a multimodal text-image understanding scenario. This provides strong evidence that MRE facilitates mutual gains across three interrelated tasks, confirming its generalizability beyond the textual domain.

\end{abstract}

\ccsdesc[500]{Do Not Use This Code~Generate the Correct Terms for Your Paper}
\ccsdesc[300]{Do Not Use This Code~Generate the Correct Terms for Your Paper}
\ccsdesc{Do Not Use This Code~Generate the Correct Terms for Your Paper}
\ccsdesc[100]{Do Not Use This Code~Generate the Correct Terms for Your Paper}

\keywords{Do, Not, Us, This, Code, Put, the, Correct, Terms, for,
  Your, Paper}


\maketitle

\section{Introduction}

Information extraction (IE)~\citep{cowie1996information, nasar2018information, xu2024large} has traditionally been a fundamental task in natural language processing, aiming to extract specific fields from unstructured text and classify them into predefined categories, ultimately transforming the raw text into structured data. Representative IE tasks include named entity recognition (NER)~\citep{nadeau2007survey}, relation extraction (RE)~\citep{zhao2024comprehensive}, and event extraction (EE)~\citep{xiang2019survey}, all of which focus on identifying meaningful words or spans in the text and assigning them to categories such as person, organization, or location. With the rapid development of multimodal models in recent years~\cite{liu_visual_2023, chen_internvl_2024, wu_deepseek-vl2_2024, lu_deepseek-vl_2024, openai_gpt-4o_2024}, multimodal information extraction \citep{gong2017multimodal, rahman2020integrating, sun2024umie} has increasingly attracted attention from the research community. Unlike text-only IE, which classifies extracted entities solely based on textual context, multimodal IE further incorporates visual information into the classification process. In other words, the model must match the extracted words or entities to corresponding image segments from the input image.

\begin{figure}[!t]
\centering
\includegraphics[width=238 pt]{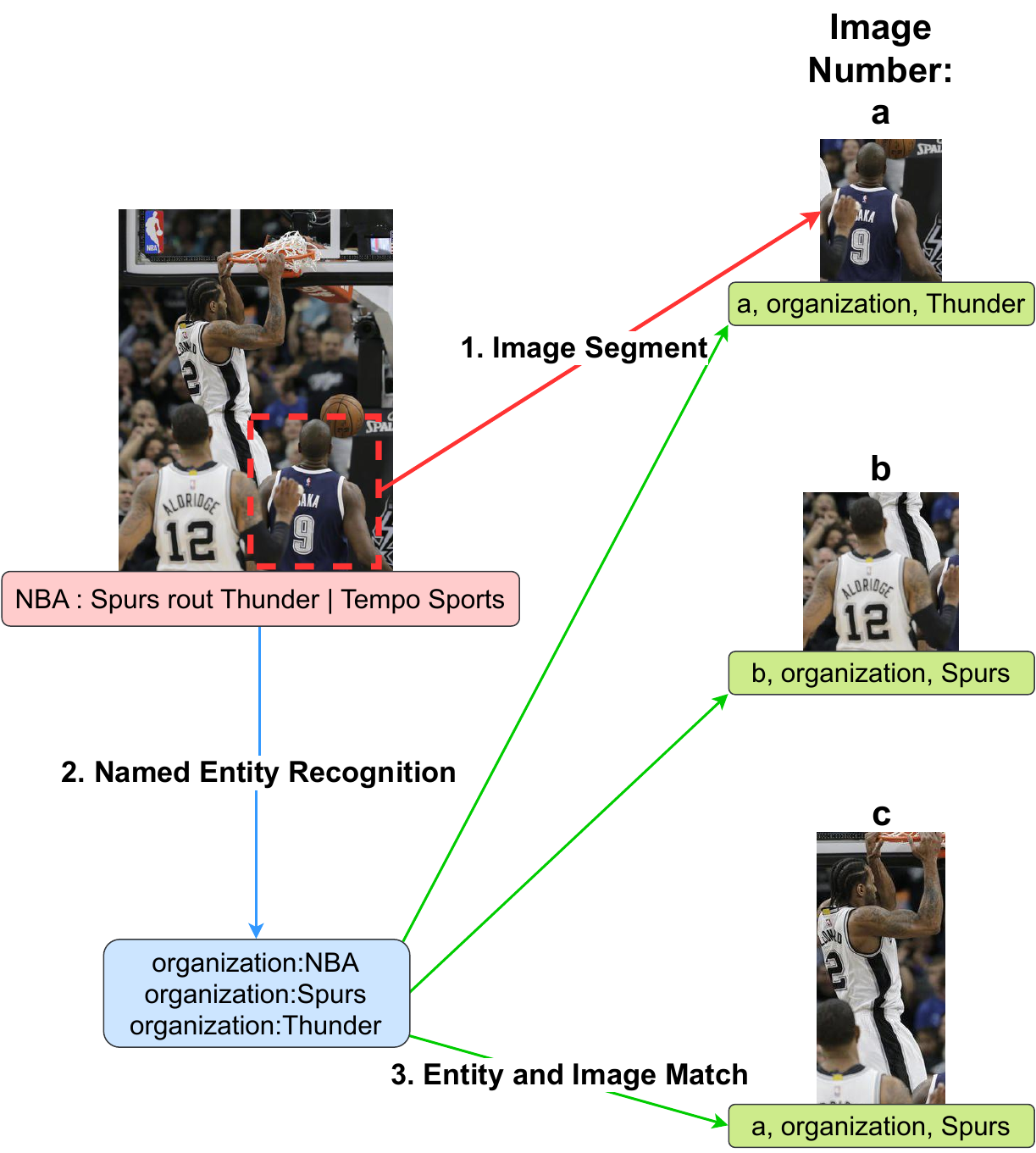}
\caption{\label{1figure1}Multimodal named entity recognition task. Contains three subtasks. Named entity recognition for text. Recognition segmentation of images. Segmentation of images and entities matching.}\Description{Multimodal named entity recognition task with three subtasks: text recognition, image segmentation, and matching entities with image segments.}

\end{figure}

As shown in \autoref{1figure1}, this is an example of a multimodal named entity recognition (MNER) task. The input consists of a raw image in the upper left corner and its accompanying textual description, typically a news headline. The model is required to perform three sub-tasks jointly. The first task is to conduct NER on the input text, extracting relevant entities and classifying them into specific labels (e.g., \texttt{organization}). The second task is to detect objects in the image and segment them into individual image patches. For instance, in \autoref{1figure1}, three human figures in the original image are detected and segmented into three patches labeled \texttt{a}, \texttt{b}, and \texttt{c}. The third task is to match the extracted textual entities with the corresponding image patches. In the figure, each person is correctly matched with their associated organization entity. This defines the multimodal NER task, which serves as the core focus of this paper. It is important to note that the original MNER task only involves fine-grained predictions and lacks any form of coarse-grained task. To incorporate and demonstrate the Mutual Reinforcement Effect (MRE)~\cite{gan2023sentence} within the MNER task, it is necessary to introduce a multimodal coarse-grained task and combine it with the fine-grained task to form a complete MMRE dataset.

\begin{figure*}[!t]
\centering
\includegraphics[width=480 pt]{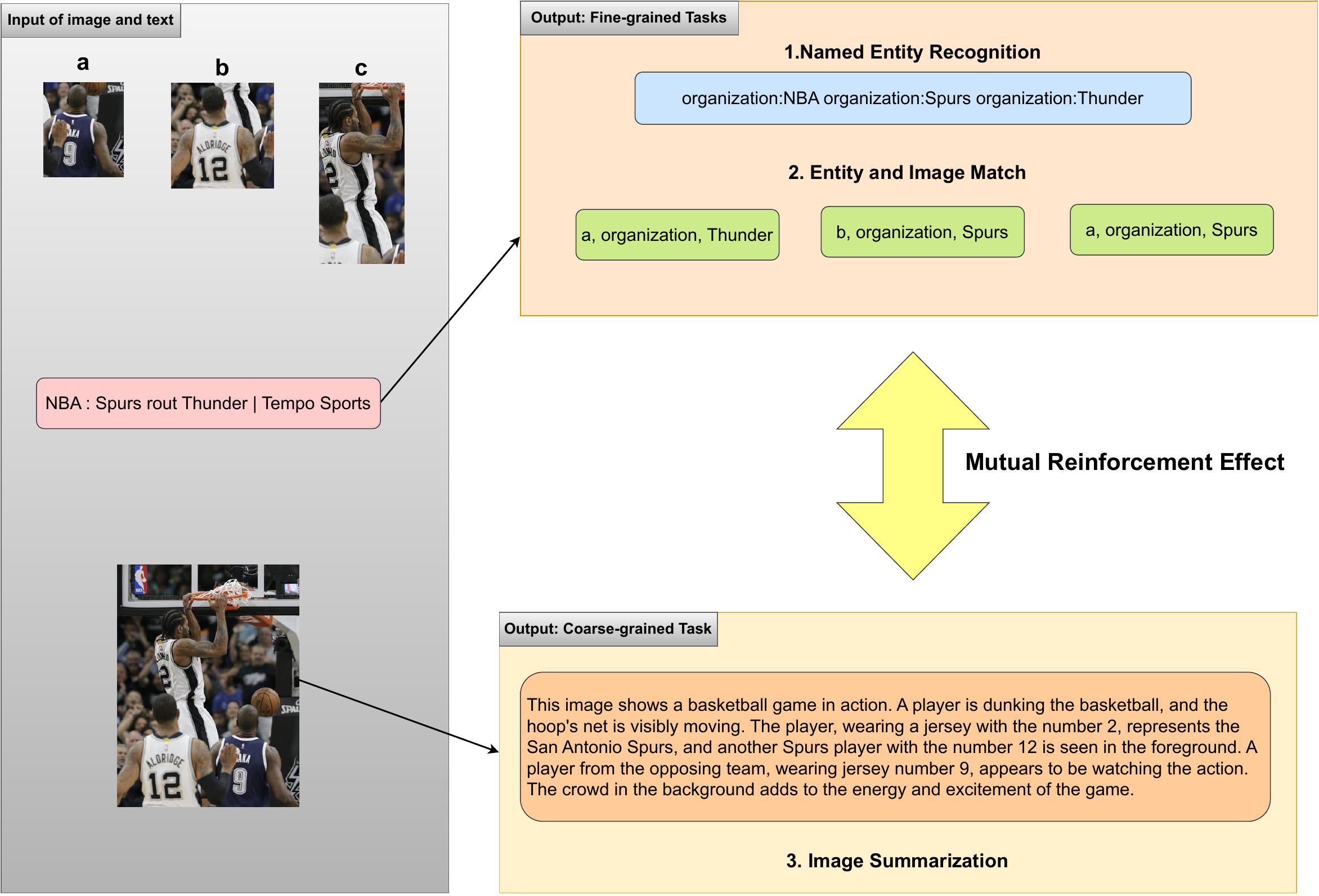}
\caption{\label{2figure2}Multimodal Mutual Reinforcement Effect Task illustration.}\Description{Multimodal Mutual Reinforcement Effect Task illustration.}

\end{figure*}

Before formally introducing the MMRE task, we briefly explain the concept of MRE in the textual domain. In text-based IE, tasks are generally divided into two levels: word-level and text-level. For example, consider the sentence: ``The food at this restaurant is delicious.'' The text-level task assigns a sentiment polarity label such as \texttt{positive}. Meanwhile, the word-level task might output a label-entity pair like \texttt{positive: delicious}. From a human perspective, understanding a sentence often involves inferring that the presence of positive words suggests an overall positive sentiment, and vice versa: knowing the sentence expresses a positive sentiment helps identify which words convey that positivity. Thus, combining the two levels allows the model to understand the input from multiple perspectives. The comprehension of one task reinforces the performance of the other. This is the essence of MRE in information extraction: joint modeling of tasks at different granularities yields better performance than modeling them separately. This practical utility of MRE is what we aim to explore in the multimodal setting.

We now formally introduce the concept of incorporating MRE into the multimodal NER task by proposing the Multimodal Mutual Reinforcement Effect task (M-MRE). As discussed earlier, the standard MNER task operates at a fine-grained level. To construct an MRE setting, we introduce an associated coarse-grained task to jointly model with MNER, enabling us to observe and verify the existence of MRE in multimodal scenarios.

As illustrated in \autoref{2figure2}, we augment the original MNER task with an image captioning task, which serves as the coarse-grained component in the multimodal setting. In the upper part of the figure, the model first performs named entity recognition on the input text, segments the image into object patches, and matches textual entity spans to corresponding image regions. In the lower part of the figure, the model generates a holistic caption for the entire image. This caption typically includes rich visual details such as team names, jersey numbers, and other contextual cues. Importantly, the fine-grained and coarse-grained tasks are inherently related: knowledge from one can enhance the other, thereby creating mutual reinforcement.

This mutual reinforcement enables performance gains across tasks. For example, a deeper understanding of the image content required for summarization can aid the model in grounding entity recognition decisions in visual context. Conversely, recognizing key textual entities can help focus the captioning task on the most salient aspects of the image.

It is worth noting that the MNER component of the M-MRE task is not identical to the original formulation. For the purpose of isolating and analyzing MRE within multimodal models, we simplify the MNER task by removing the object detection and segmentation step. Instead, the model is directly provided with candidate image patches and is only required to match them with the extracted textual entities. This design choice allows M-MRE to emphasize the interaction between language and vision-language understanding, rather than visual processing itself. In doing so, we control for confounding factors and focus specifically on the emergence of MRE within multimodal learning.

To reduce annotation costs in constructing the M-MRE task, we leverage the existing \textit{Grounded Multimodal Named Entity Recognition} (GMNER) dataset\citep{yu-etal-2023-grounded}. GMNER already contains original images, segmented image patches derived from those images, textual content, and corresponding label-entity pairs. Based on this dataset, we generate image captions for the coarse-grained task using a large language model (LLM). These captions are then manually reviewed and refined to ensure quality, resulting in the finalized M-MRE dataset.

To effectively process the M-MRE task, we further propose a \textit{Prompt Format Adapter} (PFA), a prompt-based framework designed to be fully compatible with any large vision-language model (LVLM). PFA provides a unified input-output interface that enables simultaneous handling of the three sub-tasks within M-MRE.

Experimental results demonstrate that MRE can indeed be observed in multimodal information extraction tasks, validating the utility of our proposed approach. This work opens up a novel subfield for future research in multimodal information extraction.

\noindent\textbf{Our main contributions are summarized as follows:}
\begin{itemize}
    \item We introduce the Mutual Reinforcement Effect (MRE) into multimodal information extraction for the first time, by proposing the M-MRE task and constructing a corresponding dataset.
    \item We design the Prompt Format Adapter (PFA), a unified prompt framework that enables simultaneous processing of the three sub-tasks in M-MRE with any LVLM.
    \item Through comprehensive experiments and ablation studies, we demonstrate the effectiveness of PFA and provide empirical evidence for the existence of MRE in the multimodal setting.
\end{itemize}

\section{Related Work}

\subsection{Mutlimodal Information Extraction}

\citet{yu-etal-2023-grounded} introduce the Grounded Multimodal Named Entity Recognition (GMNER) task, which requires extracting entity-type-region triples from text-image pairs on social media, and propose a hierarchical index generation framework (H-Index) that outperforms existing baselines on a newly annotated dataset. \citet{10.1145/3581783.3612322} propose the Fine-grained Multimodal Named Entity Recognition and Grounding (FMNERG) task, which aims to extract entity-type-object triples with fine-grained categories, and introduce a T5-based generative framework (TIGER) that achieves state-of-the-art results on a newly constructed Twitter dataset. And \citet{10.1145/3664647.3681598} present a Generative Multimodal Data Augmentation (GMDA) framework for low-resource MNER, combining label-aware text generation with image synthesis to improve entity recognition performance under both full and limited supervision. \citet{liu-etal-2019-graph} propose a graph convolution-based model for information extraction from visually rich documents (VRDs), effectively integrating textual and visual layout features to outperform BiLSTM-CRF baselines on real-world datasets. There are also a number of other multimodal IE \citet{dong-etal-2020-multi-modal} subtasks that encompass relationship extraction \citep{zheng2021multimodal}, event extraction \citep{zhang2017improving}, sentiment analysis \citep{gandhi2023multimodal}, and so on.

\subsection{Mutual Reinforcement Effect}

Mutual Reinforcement Effects (MRE) was first proposed in the field of text information extraction. \citet{gan2023sentence} propose the Sentence-to-Label Generation (SLG) framework for multi-task learning of Japanese Sentence Classification (SC) and Named Entity Recognition (NER), unifying both tasks under a generative paradigm. They construct a new SCNM dataset by annotating sentence-level categories on an existing Japanese Wikipedia NER corpus and demonstrate that jointly learning SC and NER tasks can yield MRE, where the performance of each task benefits from the presence of the other. Their results show that SLG improves SC and NER accuracy by 1.13 and 1.06 points, respectively, over single-task baselines. They also introduce a Constraint Mechanism (CM) to ensure output format correctness and perform incremental learning with the Shinra NER corpus to improve the model’s NER capability. Subsequent studies have extended MRE to other languages and various sub-tasks within information extraction, often in conjunction with large language models (LLMs) under specialized training paradigms \citep{gan2024mmm}. These MRE-based approaches consistently outperform conventional training methods that treat each IE task in isolation.

Although MRE remains a relatively underexplored research area, its formulation—combining fine-grained and coarse-grained tasks—is especially well-suited for multimodal information extraction. This is because text and images naturally complement and reinforce one another, making mutual enhancement not only possible but also beneficial. This observation serves as the primary motivation behind our work.

\section{Multimodal Mutual Reinforcement Effect}

\begin{figure*}[!t]
\centering
\includegraphics[width=480 pt]{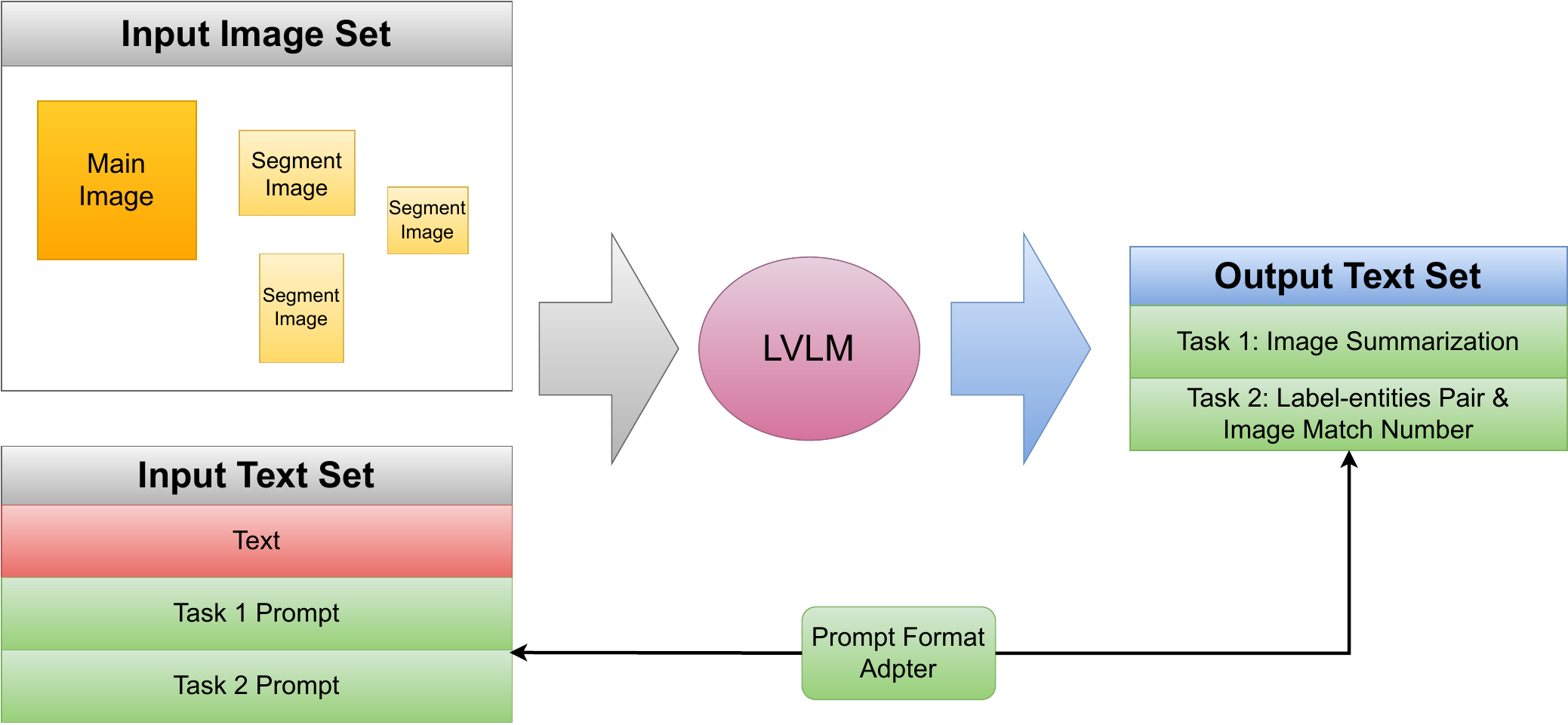}
\caption{\label{3figure3}Inputs and outputs of Large Vision-Language Models while processing M-MRE tasks.}\Description{Inputs and outputs of Large Vision-Language Models while processing M-MRE tasks.}

\end{figure*}

As previously outlined, the M-MRE task comprises both fine-grained and coarse-grained subtasks. In this section, we first describe the overall processing pipeline for handling the M-MRE task, followed by an explanation of how the M-MRE dataset is constructed. We then provide a detailed description of the proposed Prompt Format Adapter (PFA).

As illustrated in \autoref{3figure3}, the input and output structure of large vision-language models (LVLMs) when processing the M-MRE task consists of two main components: an image group and a text group. The image group includes the main image and several segmented image patches derived from it. The text group contains the accompanying textual input, which serves as the target text for named entity recognition (NER). Within the text group, the green-highlighted sections represent the Prompt Format Adapter (PFA), which unifies both input and output formats, thereby simplifying the LVLM's processing of multiple sub-tasks.

The input-side PFA is composed of two prompt instructions corresponding to the two subtasks: Task~1 (image summarization) and Task~2 (NER and segment-image matching). It is important to note that Task~2 is not separated into two independent components. This is because the segment-image matching inherently involves identifying and labeling entity spans; that is, it subsumes the label-entity pair extraction from NER. As a result, we merge the prompts for these two aspects into a single instruction to guide the model's response generation effectively.

Once all input images and textual prompts are prepared, they are sequentially passed into the LVLM. The model then generates structured outputs for both tasks. Crucially, the two green text prompts in the input are formatted via the PFA, which also ensures that the output conforms to the structural requirements of information extraction—specifically, producing data that can be directly stored and reused as structured text.

For Task~1, the output consists of a natural language summary that describes the visual content of the image. For Task~2, the output is a sequence of label-entity pairs in the form of tuples, organized according to their extraction order. Following this, the model generates a second sequence of tuples, each pairing a segmented image number with the corresponding entity span, thereby completing the entity-to-image matching task.

A detailed explanation of the PFA’s design will be provided in the subsequent section after we describe the construction process of the M-MRE dataset. The above describes the complete workflow for applying LVLMs to the M-MRE task, enabling the simultaneous processing of fine-grained and coarse-grained subtasks through a unified input-output interface.

\subsection{Multimodal Mutual Reinforcement Effect Dataset}

One major challenge in validating the Mutual Reinforcement Effect (MRE) in multimodal information extraction is the lack of a dedicated multimodal MRE dataset. To address this gap, we construct the M-MRE dataset by augmenting an existing multimodal NER dataset with additional coarse-grained annotations using a large language model (LLM). Specifically, we utilize the \textit{Grounded Multimodal Named Entity Recognition} (GMNER) dataset, which already includes fine-grained multimodal NER labels. This enables us to build the M-MRE dataset through incremental annotation.

\begin{figure}[!ht]
\centering
\includegraphics[width=238 pt]{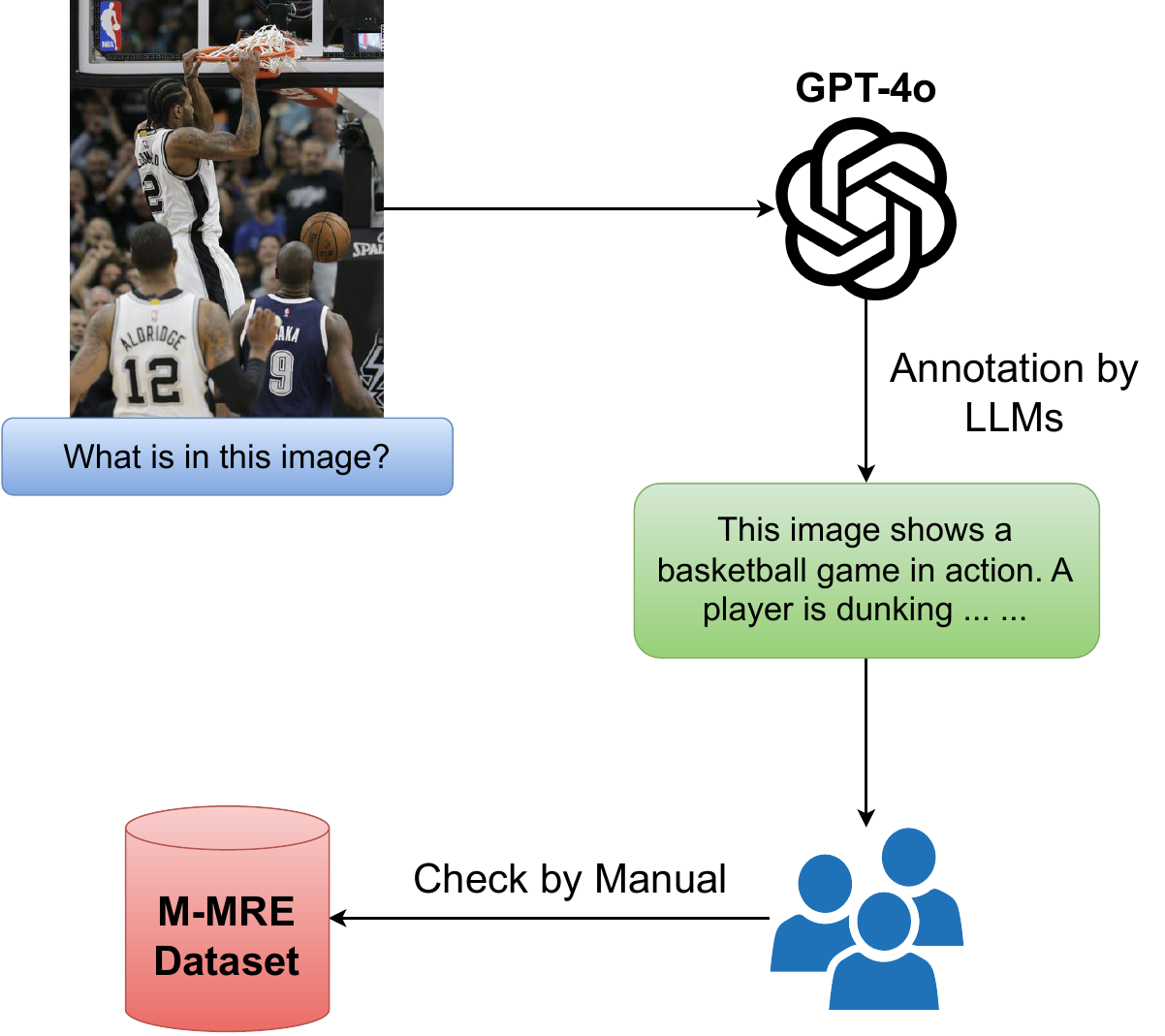}
\caption{\label{4figure4}Inputs and outputs of Large Vision-Language Models while processing M-MRE tasks.}\Description{Inputs and outputs of Large Vision-Language Models while processing M-MRE tasks.}

\end{figure}

The GMNER dataset itself is built upon two popular multimodal NER datasets, \citet{zhang2018adaptive} and \citet{yu-etal-2020-improving-multimodal}. GMNER introduces object-level segmentation for the original images and aligns the segmented patches with corresponding entity spans. In our work, we further augment the GMNER dataset by generating image-level summarizations, thereby forming the basis for the coarse-grained component required in the M-MRE task.

\begin{figure*}[!ht]
\centering
\includegraphics[width=480 pt]{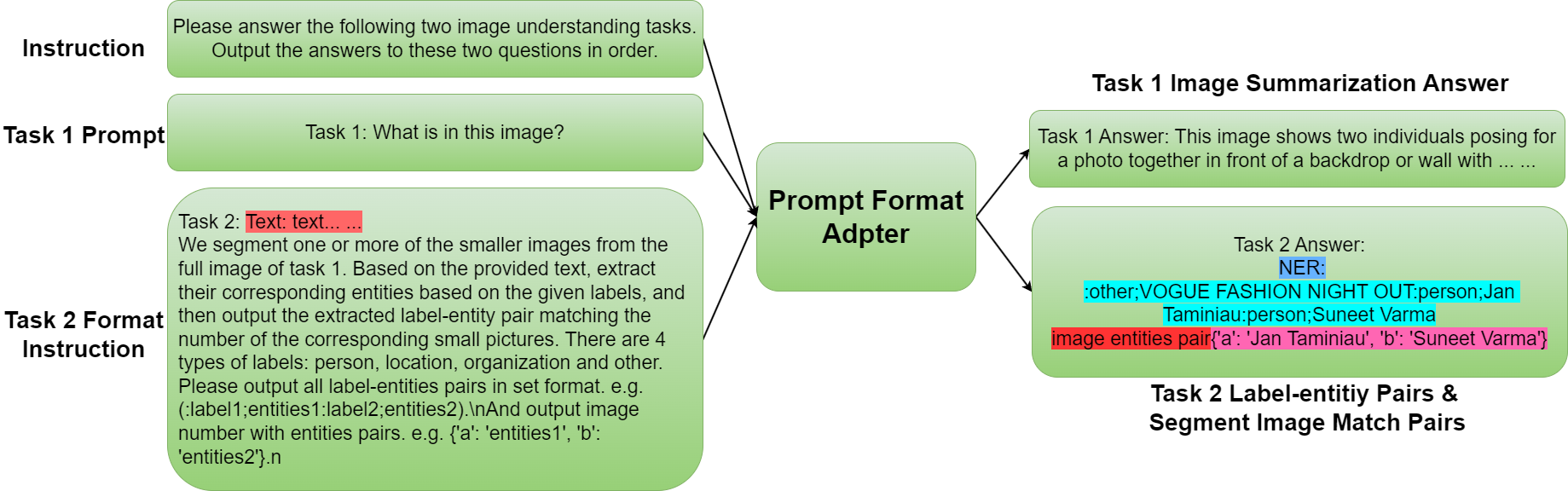}
\caption{\label{5figure5}Inputs and outputs of Large Vision-Language Models while processing M-MRE tasks.}\Description{Inputs and outputs of Large Vision-Language Models while processing M-MRE tasks.}

\end{figure*}

As illustrated in \autoref{4figure4}, we input both the main image and a predefined instruction prompt template into the GPT-4o model, specifically the \texttt{yu-etal-2023-grounded} version. The model then generates a descriptive summary of the image content. To ensure quality and consistency, each generated summary undergoes cross-checking and refinement by multiple annotators. The resulting M-MRE dataset thus combines both fine-grained and coarse-grained annotations.

\begin{table}[!t]
\centering
\caption{Statistics of Task 1 image summarization answers in the M-MRE dataset.}
\label{table1}
\begin{tabular}{lcccc}
\toprule[0.4mm]
\rowcolor{mygray} \textbf{Metric} & \textbf{Max} & \textbf{Min} & \textbf{Median} & \textbf{Mean} \\ \hline \hline
Character Count & 573 & 40 & 317 & 318 \\
Word Count      & 97  & 10 & 54  & 53 \\
\bottomrule[0.4mm]
\end{tabular}
\end{table}

Annotation statistics for the image summarization component are shown in \autoref{table1}. The average number of words per summary is 53, with a median of 54. These results indicate that the GPT-4o model produces stable and well-balanced outputs—neither too short to omit essential image details, nor excessively long to the point of over-describing the visual content.

\subsection{Prompt Format Adapter}

We now present the design of the \textit{Prompt Format Adapter} (PFA), which plays two key roles in the M-MRE framework. First, it serves as a prompt to guide the model’s output behavior. Second, it enforces a consistent output format, enabling the generation of structured and machine-readable text data.

As shown in \autoref{5figure5}, PFA consists of two components: input formatting and output formatting. The input part includes three prompt modules. The first module is a general instruction that tells the model to answer the following two tasks while suppressing unnecessary verbose outputs. This instruction is crucial in LVLMs, which are typically trained via instruction tuning during post-training and respond effectively to such guidance.

The second and third modules correspond to \textbf{Task 1} and \textbf{Task 2}, respectively. Among these, the instruction format for Task 2 is more complex, as it involves two sub-tasks. For the NER sub-task, we define the output format as \texttt{:label;entities}. Inspired by prior work on textual MRE-based information extraction, we adopt the colon (\texttt{:}) to mark the beginning of a label, followed by a semicolon (\texttt{;}) to separate the label from its corresponding entity span. The entity text is then appended directly. This format is easily extendable to multiple label-entity pairs in sequence (e.g., \texttt{:label1;entities1:label2;entities2:label3;entities3}~$\ldots$), allowing a compact yet structured representation.

For the second sub-task of Task 2, which matches segmented image patches with entities, we define the output format as a set of key-value mappings, such as \texttt{\{'a': 'entities1'\}}, where each patch ID corresponds to the entity it represents. This standardization enables reliable and scalable parsing of model outputs.

On the output side, PFA further enforces fixed textual markers to clearly indicate the beginning of responses for each task. Specifically, \texttt{Task 1 Answer:} and \texttt{Task 2 Answer:} are used as headers for the model’s output. Within the Task 2 response, we include an additional level of guidance to distinguish the two sub-tasks: the \texttt{NER:} prefix (highlighted in green in \autoref{5figure5}) denotes the NER sub-task output, while the \texttt{image-entity pair:} prefix (highlighted in red) denotes the output for image-entity match.

In summary, all sub-tasks in the M-MRE framework are prompted and output in a unified, structured format. Since PFA only modifies and constrains the textual input and output, without altering the model architecture or weights, it remains model-agnostic and can be seamlessly applied to any LVLM in a plug-and-play manner.

\section{Experiment Setup}

For the base model, we select \texttt{Qwen2.5-VL-7B-Instruct}\footnote{\raggedright\url{https://huggingface.co/Qwen/Qwen2.5-VL-7B-Instruct}}, a state-of-the-art open-source LVLM known for its strong performance in multimodal understanding tasks. We use the finalized M-MRE dataset, which contains 5,533 annotated samples, and randomly split it into training and testing sets with a 7/3 ratio.

We train the model using full-parameter tuning with \texttt{bfloat16} (BF16) precision. The training hyperparameters are as follows: a batch size of 4, a learning rate of $1\text{e}{-5}$ for the text encoder, $2\text{e}{-6}$ for the vision encoder, and a unified learning rate of $2\text{e}{-6}$ for joint training. The model is trained for 3 epochs.

We adopt the Qwen fine-tuning framework\footnote{\raggedright\url{https://github.com/2U1/Qwen2-VL-Finetune}} for training. The multimodal input format adheres to the \texttt{LLaVA} input standard, ensuring compatibility with existing instruction-following LVLM pipelines. All experiments are conducted on a distributed setup with four NVIDIA A100 80GB GPUs.

\subsection{Evaluation}

Evaluating the M-MRE task is non-trivial due to its multi-task nature and diverse output structure. To facilitate evaluation, we first segment the model's output text into three components based on predefined keywords: (1) the image summarization text for Task~1, and (2) the structured text outputs for the two sub-tasks in Task~2.

For \textbf{Task 1 (image summarization)}, we adopt widely used natural language generation metrics: \textbf{ROUGE-1}, \textbf{ROUGE-2}, \textbf{ROUGE-L}, and \textbf{BLEU}. These metrics evaluate the content overlap between the generated summary and the human-annotated ground truth.

For \textbf{Task 2}, we use distinct metrics for the two sub-tasks. The \textbf{NER sub-task} is evaluated using the \textbf{F1 score}, which measures the harmonic mean of precision and recall over extracted label-entity pairs. The \textbf{image-entity matching sub-task} is evaluated using \textbf{accuracy}, which measures the proportion of correctly matched image patches to their corresponding entities.

The F1 score is computed as follows:
\begin{equation}
\text{F1 Score} = \frac{2 \cdot \text{Precision} \cdot \text{Recall}}{\text{Precision} + \text{Recall}}
\end{equation}

\vspace{0.5em}

\begin{algorithm}[h]
\caption{Evaluation Procedure for M-MRE Tasks}
\label{evaluation}
\KwIn{Predicted text $T_{\text{pred}}$, Reference text $T_{\text{ref}}$}
\KwOut{BLEU, ROUGE-1/2/L, F1\_NER, ImageEntity\_Accuracy}

Extract summary texts from $T_{\text{pred}}$ and $T_{\text{ref}}$ using regex pattern ``\texttt{Task 1 Answer:}''\;
Extract NER text using pattern ``\texttt{NER::}''\;
Extract image-entity pairs using pattern ``\texttt{image entities pair}''\;

Parse NER label-entity pairs using delimiter \texttt{:} and \texttt{;}\;
Convert image-entity mappings to dictionaries\;

Compute ROUGE and BLEU between predicted and reference summaries\;
Match NER label-entity pairs and compute precision and recall\;
Compute F1 score: $F1 = \frac{2PR}{P+R}$\;

Match predicted and reference image-entity pairs and compute accuracy\;

\Return All metrics: \{BLEU, ROUGE-1/2/L, F1\_NER, ImageEntity\_Accuracy\}\;
\end{algorithm}

\noindent The evaluation pipeline is illustrated in the pseudocode of \textbf{\autoref{evaluation}}. In our method, word-level predictions are first parsed into individual label-entity pairs, such as \texttt{:neutral;unique}, \texttt{:positive;boring}, \texttt{:positive;sad}. These predicted pairs are then matched against the ground-truth label-entity pairs.

Precision is defined as the number of correctly matched pairs divided by the number of predicted pairs. Recall is computed as the number of correctly matched pairs divided by the number of ground-truth pairs. This matching-based evaluation provides a fine-grained and realistic measurement of model performance in structured prediction tasks.

\section{Results}

\begin{table}[h]
\centering
\caption{Main evaluation results on the M-MRE task using PFA.}
\label{tab:main_results}
\begin{tabular}{lccc}
\toprule[0.4mm]
\rowcolor{mygray} \textbf{Method} & \textbf{F1} & \textbf{Accuracy} & \textbf{BLEU} \\ \hline \hline
Single TS    & -     & -     & 15.65 \\
Single MNER  & \textbf{86.70} & \textbf{90.05} & -     \\
M-MRE (PFA)  & 84.96 & 88.83 & \textbf{15.81} \\
\midrule
\rowcolor{mygray} \textbf{Method} & \textbf{ROUGE-1} & \textbf{ROUGE-2} & \textbf{ROUGE-L} \\
\midrule
Single TS    & 57.22 & 28.41 & 42.22 \\
Single MNER  & -     & -     & -     \\
M-MRE (PFA)  & \textbf{57.30} & \textbf{28.72} & \textbf{42.64} \\
\bottomrule[0.4mm]
\end{tabular}
\end{table}

As shown in \autoref{tab:main_results}, we present the complete evaluation results of the M-MRE task using our proposed PFA framework. Here, \textbf{Single TS} denotes performing only the image-to-text summarization task, representing a coarse-grained multimodal task. \textbf{Single MNER} refers to performing only the MNER task, with both sub-tasks evaluated, representing a fine-grained multimodal task. \textbf{M-MRE} indicates the joint execution of both granularities via the full MRE setup.

We observe that the performance of the image summarization task improves across all four metrics (BLEU, ROUGE-1/2/L) in the M-MRE setting compared to the Single TS baseline. This indicates that the fine-grained task provides helpful signals for understanding the image at a more holistic level. On the other hand, the performance of MNER slightly drops in the M-MRE setting. As shown in our later ablation study, this degradation may be attributed to the PFA design for MNER, rather than an indication that coarse-grained tasks fail to reinforce fine-grained tasks. We investigate this further in the next subsection.

\section{Ablation Study of Multimodal MRE}

\begin{table}[ht]
\centering
\caption{Ablation study results without PFA (simplified input/output).}
\label{tab:ablation}
\begin{tabular}{lccc}
\toprule[0.4mm]
\rowcolor{mygray} \textbf{Method} & \textbf{F1} & \textbf{Accuracy} & \textbf{BLEU} \\ \hline\hline
Single TS    & -     & -     & 14.25 \\
Single MNER  & 84.21 & 86.79 & -     \\
M-MRE (no PFA) & 83.91 & \textbf{87.50} & \textbf{15.61} \\
\midrule
\rowcolor{mygray}\textbf{Method} & \textbf{ROUGE-1} & \textbf{ROUGE-2} & \textbf{ROUGE-L} \\
\midrule
Single TS    & 51.24 & 25.45 & 38.09 \\
Single MNER  & -     & -     & -     \\
M-MRE (no PFA) & \textbf{56.96} & \textbf{28.71} & \textbf{42.44} \\
\bottomrule[0.4mm]
\end{tabular}
\end{table}

To further isolate the effect of MRE and eliminate other influencing factors, we conduct an ablation experiment by simplifying both the input and output format. Specifically, we remove PFA and directly input only the original image and its associated text. The output contains only the image summary and the MNER predictions: label-entity pairs and segment-image matches.

Using the same experimental settings, the results are reported in \autoref{tab:ablation}. We observe a noticeable drop in both sub-tasks of the MNER component compared to the PFA-based setting, which strongly supports the effectiveness of the PFA framework. Furthermore, the accuracy of the image-entity matching sub-task significantly improves when jointly modeled with image summarization, compared to the single MNER baseline. Although the F1 score for label-entity pairs still does not surpass the Single MNER baseline, the gap is reduced by nearly half. Meanwhile, the performance of image summarization remains consistent with previous results, confirming its robustness.

Overall, this ablation study not only verifies the utility of PFA but also provides additional evidence for the presence of the Mutual Reinforcement Effect in multimodal information extraction, as captured by our M-MRE dataset.

\section{Conclusion}

In this work, we introduce \textbf{M-MRE}, a novel multimodal information extraction task designed to investigate the \textit{Mutual Reinforcement Effect} (MRE) between coarse-grained and fine-grained subtasks in a multimodal setting. To support this task, we construct a high-quality dataset by extending the GMNER benchmark with image summarization annotations generated by GPT-4o and manually verified for accuracy. 

We further propose the \textbf{Prompt Format Adapter} (PFA), a lightweight yet effective prompting framework that unifies task instructions and output formatting, enabling seamless adaptation to any vision-language model (LVLM). Experimental results demonstrate that the M-MRE framework not only improves coarse-grained performance (image summarization) through fine-grained signals (MNER), but also maintains robust performance across sub-tasks. Ablation studies confirm the effectiveness of PFA and provide strong evidence that MRE exists in multimodal information extraction, especially when properly formatted prompts are used.

Our work opens up a new research direction at the intersection of multimodal learning and structured information extraction. We hope the M-MRE dataset and findings in this study will serve as a valuable resource for future studies on task interaction, prompt design, and model interpretability in LVLMs.

\begin{acks}
To Robert, for the bagels and explaining CMYK and color spaces.
\end{acks}

\bibliographystyle{ACM-Reference-Format}
\bibliography{sample-base}


\begin{thebibliography}{26}


\ifx \showCODEN    \undefined \def \showCODEN     #1{\unskip}     \fi
\ifx \showISBNx    \undefined \def \showISBNx     #1{\unskip}     \fi
\ifx \showISBNxiii \undefined \def \showISBNxiii  #1{\unskip}     \fi
\ifx \showISSN     \undefined \def \showISSN      #1{\unskip}     \fi
\ifx \showLCCN     \undefined \def \showLCCN      #1{\unskip}     \fi
\ifx \shownote     \undefined \def \shownote      #1{#1}          \fi
\ifx \showarticletitle \undefined \def \showarticletitle #1{#1}   \fi
\ifx \showURL      \undefined \def \showURL       {\relax}        \fi
\providecommand\bibfield[2]{#2}
\providecommand\bibinfo[2]{#2}
\providecommand\natexlab[1]{#1}
\providecommand\showeprint[2][]{arXiv:#2}

\bibitem[Chen et~al\mbox{.}(2024)]%
        {chen_internvl_2024}
\bibfield{author}{\bibinfo{person}{Zhe Chen}, \bibinfo{person}{Jiannan Wu}, \bibinfo{person}{Wenhai Wang}, \bibinfo{person}{Weijie Su}, \bibinfo{person}{Guo Chen}, \bibinfo{person}{Sen Xing}, \bibinfo{person}{Muyan Zhong}, \bibinfo{person}{Qinglong Zhang}, \bibinfo{person}{Xizhou Zhu}, \bibinfo{person}{Lewei Lu}, \bibinfo{person}{Bin Li}, \bibinfo{person}{Ping Luo}, \bibinfo{person}{Tong Lu}, \bibinfo{person}{Yu Qiao}, {and} \bibinfo{person}{Jifeng Dai}.} \bibinfo{year}{2024}\natexlab{}.
\newblock \showarticletitle{{InternVL}: {Scaling} up {Vision} {Foundation} {Models} and {Aligning} for {Generic} {Visual}-{Linguistic} {Tasks}}. In \bibinfo{booktitle}{\emph{Proceedings of the {IEEE}/{CVF} {Conference} on {Computer} {Vision} and {Pattern} {Recognition}}}. \bibinfo{pages}{24185--24198}.
\newblock
\urldef\tempurl%
\url{https://openaccess.thecvf.com/content/CVPR2024/html/Chen_InternVL_Scaling_up_Vision_Foundation_Models_and_Aligning_for_Generic_CVPR_2024_paper.html}
\showURL{%
\tempurl}


\bibitem[Cowie and Lehnert(1996)]%
        {cowie1996information}
\bibfield{author}{\bibinfo{person}{Jim Cowie} {and} \bibinfo{person}{Wendy Lehnert}.} \bibinfo{year}{1996}\natexlab{}.
\newblock \showarticletitle{Information extraction}.
\newblock \bibinfo{journal}{\emph{Commun. ACM}} \bibinfo{volume}{39}, \bibinfo{number}{1} (\bibinfo{year}{1996}), \bibinfo{pages}{80--91}.
\newblock


\bibitem[Dong et~al\mbox{.}(2020)]%
        {dong-etal-2020-multi-modal}
\bibfield{author}{\bibinfo{person}{Xin~Luna Dong}, \bibinfo{person}{Hannaneh Hajishirzi}, \bibinfo{person}{Colin Lockard}, {and} \bibinfo{person}{Prashant Shiralkar}.} \bibinfo{year}{2020}\natexlab{}.
\newblock \showarticletitle{Multi-modal Information Extraction from Text, Semi-structured, and Tabular Data on the Web}. In \bibinfo{booktitle}{\emph{Proceedings of the 58th Annual Meeting of the Association for Computational Linguistics: Tutorial Abstracts}}, \bibfield{editor}{\bibinfo{person}{Agata Savary} {and} \bibinfo{person}{Yue Zhang}} (Eds.). \bibinfo{publisher}{Association for Computational Linguistics}, \bibinfo{address}{Online}, \bibinfo{pages}{23--26}.
\newblock
\href{https://doi.org/10.18653/v1/2020.acl-tutorials.6}{doi:\nolinkurl{10.18653/v1/2020.acl-tutorials.6}}


\bibitem[Gan et~al\mbox{.}(2024)]%
        {gan2024mmm}
\bibfield{author}{\bibinfo{person}{Chengguang Gan}, \bibinfo{person}{Sunbowen Lee}, \bibinfo{person}{Qingyu Yin}, \bibinfo{person}{Xinyang He}, \bibinfo{person}{Hanjun Wei}, \bibinfo{person}{Yunhao Liang}, \bibinfo{person}{Younghun Lim}, \bibinfo{person}{Shijian Wang}, \bibinfo{person}{Hexiang Huang}, \bibinfo{person}{Qinghao Zhang}, {et~al\mbox{.}}} \bibinfo{year}{2024}\natexlab{}.
\newblock \showarticletitle{Mmm: Multilingual mutual reinforcement effect mix datasets \& test with open-domain information extraction large language models}.
\newblock \bibinfo{journal}{\emph{arXiv preprint arXiv:2407.10953}} (\bibinfo{year}{2024}).
\newblock


\bibitem[Gan et~al\mbox{.}(2023)]%
        {gan2023sentence}
\bibfield{author}{\bibinfo{person}{Chengguang Gan}, \bibinfo{person}{Qinghao Zhang}, {and} \bibinfo{person}{Tatsunori Mori}.} \bibinfo{year}{2023}\natexlab{}.
\newblock \showarticletitle{Sentence-to-label generation framework for multi-task learning of japanese sentence classification and named entity recognition}. In \bibinfo{booktitle}{\emph{International Conference on Applications of Natural Language to Information Systems}}. Springer, \bibinfo{pages}{257--270}.
\newblock


\bibitem[Gandhi et~al\mbox{.}(2023)]%
        {gandhi2023multimodal}
\bibfield{author}{\bibinfo{person}{Ankita Gandhi}, \bibinfo{person}{Kinjal Adhvaryu}, \bibinfo{person}{Soujanya Poria}, \bibinfo{person}{Erik Cambria}, {and} \bibinfo{person}{Amir Hussain}.} \bibinfo{year}{2023}\natexlab{}.
\newblock \showarticletitle{Multimodal sentiment analysis: A systematic review of history, datasets, multimodal fusion methods, applications, challenges and future directions}.
\newblock \bibinfo{journal}{\emph{Information Fusion}}  \bibinfo{volume}{91} (\bibinfo{year}{2023}), \bibinfo{pages}{424--444}.
\newblock


\bibitem[Gong et~al\mbox{.}(2017)]%
        {gong2017multimodal}
\bibfield{author}{\bibinfo{person}{Dihong Gong}, \bibinfo{person}{Daisy~Zhe Wang}, {and} \bibinfo{person}{Yang Peng}.} \bibinfo{year}{2017}\natexlab{}.
\newblock \showarticletitle{Multimodal learning for web information extraction}. In \bibinfo{booktitle}{\emph{Proceedings of the 25th ACM international conference on Multimedia}}. \bibinfo{pages}{288--296}.
\newblock


\bibitem[Li et~al\mbox{.}(2024)]%
        {10.1145/3664647.3681598}
\bibfield{author}{\bibinfo{person}{Ziyan Li}, \bibinfo{person}{Jianfei Yu}, \bibinfo{person}{Jia Yang}, \bibinfo{person}{Wenya Wang}, \bibinfo{person}{Li Yang}, {and} \bibinfo{person}{Rui Xia}.} \bibinfo{year}{2024}\natexlab{}.
\newblock \showarticletitle{Generative Multimodal Data Augmentation for Low-Resource Multimodal Named Entity Recognition}. In \bibinfo{booktitle}{\emph{Proceedings of the 32nd ACM International Conference on Multimedia}} (Melbourne VIC, Australia) \emph{(\bibinfo{series}{MM '24})}. \bibinfo{publisher}{Association for Computing Machinery}, \bibinfo{address}{New York, NY, USA}, \bibinfo{pages}{7336–7345}.
\newblock
\showISBNx{9798400706868}
\href{https://doi.org/10.1145/3664647.3681598}{doi:\nolinkurl{10.1145/3664647.3681598}}


\bibitem[Liu et~al\mbox{.}(2023)]%
        {liu_visual_2023}
\bibfield{author}{\bibinfo{person}{Haotian Liu}, \bibinfo{person}{Chunyuan Li}, \bibinfo{person}{Qingyang Wu}, {and} \bibinfo{person}{Yong~Jae Lee}.} \bibinfo{year}{2023}\natexlab{}.
\newblock \bibinfo{title}{Visual {Instruction} {Tuning}}.
\newblock
\href{https://doi.org/10.48550/arXiv.2304.08485}{doi:\nolinkurl{10.48550/arXiv.2304.08485}}
\newblock
\shownote{arXiv:2304.08485 [cs]}.


\bibitem[Liu et~al\mbox{.}(2019)]%
        {liu-etal-2019-graph}
\bibfield{author}{\bibinfo{person}{Xiaojing Liu}, \bibinfo{person}{Feiyu Gao}, \bibinfo{person}{Qiong Zhang}, {and} \bibinfo{person}{Huasha Zhao}.} \bibinfo{year}{2019}\natexlab{}.
\newblock \showarticletitle{Graph Convolution for Multimodal Information Extraction from Visually Rich Documents}. In \bibinfo{booktitle}{\emph{Proceedings of the 2019 Conference of the North {A}merican Chapter of the Association for Computational Linguistics: Human Language Technologies, Volume 2 (Industry Papers)}}, \bibfield{editor}{\bibinfo{person}{Anastassia Loukina}, \bibinfo{person}{Michelle Morales}, {and} \bibinfo{person}{Rohit Kumar}} (Eds.). \bibinfo{publisher}{Association for Computational Linguistics}, \bibinfo{address}{Minneapolis, Minnesota}, \bibinfo{pages}{32--39}.
\newblock
\href{https://doi.org/10.18653/v1/N19-2005}{doi:\nolinkurl{10.18653/v1/N19-2005}}


\bibitem[Lu et~al\mbox{.}(2024)]%
        {lu_deepseek-vl_2024}
\bibfield{author}{\bibinfo{person}{Haoyu Lu}, \bibinfo{person}{Wen Liu}, \bibinfo{person}{Bo Zhang}, \bibinfo{person}{Bingxuan Wang}, \bibinfo{person}{Kai Dong}, \bibinfo{person}{Bo Liu}, \bibinfo{person}{Jingxiang Sun}, \bibinfo{person}{Tongzheng Ren}, \bibinfo{person}{Zhuoshu Li}, \bibinfo{person}{Hao Yang}, \bibinfo{person}{Yaofeng Sun}, \bibinfo{person}{Chengqi Deng}, \bibinfo{person}{Hanwei Xu}, \bibinfo{person}{Zhenda Xie}, {and} \bibinfo{person}{Chong Ruan}.} \bibinfo{year}{2024}\natexlab{}.
\newblock \bibinfo{title}{{DeepSeek}-{VL}: {Towards} {Real}-{World} {Vision}-{Language} {Understanding}}.
\newblock
\href{https://doi.org/10.48550/arXiv.2403.05525}{doi:\nolinkurl{10.48550/arXiv.2403.05525}}
\newblock
\shownote{arXiv:2403.05525 [cs]}.


\bibitem[Nadeau and Sekine(2007)]%
        {nadeau2007survey}
\bibfield{author}{\bibinfo{person}{David Nadeau} {and} \bibinfo{person}{Satoshi Sekine}.} \bibinfo{year}{2007}\natexlab{}.
\newblock \showarticletitle{A survey of named entity recognition and classification}.
\newblock \bibinfo{journal}{\emph{Lingvisticae Investigationes}} \bibinfo{volume}{30}, \bibinfo{number}{1} (\bibinfo{year}{2007}), \bibinfo{pages}{3--26}.
\newblock


\bibitem[Nasar et~al\mbox{.}(2018)]%
        {nasar2018information}
\bibfield{author}{\bibinfo{person}{Zara Nasar}, \bibinfo{person}{Syed~Waqar Jaffry}, {and} \bibinfo{person}{Muhammad~Kamran Malik}.} \bibinfo{year}{2018}\natexlab{}.
\newblock \showarticletitle{Information extraction from scientific articles: a survey}.
\newblock \bibinfo{journal}{\emph{Scientometrics}} \bibinfo{volume}{117}, \bibinfo{number}{3} (\bibinfo{year}{2018}), \bibinfo{pages}{1931--1990}.
\newblock


\bibitem[{OpenAI}(2024)]%
        {openai_gpt-4o_2024}
\bibfield{author}{\bibinfo{person}{{OpenAI}}.} \bibinfo{year}{2024}\natexlab{}.
\newblock \bibinfo{title}{{GPT}-4o {System} {Card}}.
\newblock
\urldef\tempurl%
\url{http://arxiv.org/abs/2410.21276}
\showURL{%
\tempurl}


\bibitem[Rahman et~al\mbox{.}(2020)]%
        {rahman2020integrating}
\bibfield{author}{\bibinfo{person}{Wasifur Rahman}, \bibinfo{person}{Md~Kamrul Hasan}, \bibinfo{person}{Sangwu Lee}, \bibinfo{person}{Amir Zadeh}, \bibinfo{person}{Chengfeng Mao}, \bibinfo{person}{Louis-Philippe Morency}, {and} \bibinfo{person}{Ehsan Hoque}.} \bibinfo{year}{2020}\natexlab{}.
\newblock \showarticletitle{Integrating multimodal information in large pretrained transformers}. In \bibinfo{booktitle}{\emph{Proceedings of the conference. Association for computational linguistics. Meeting}}, Vol.~\bibinfo{volume}{2020}. \bibinfo{pages}{2359}.
\newblock


\bibitem[Sun et~al\mbox{.}(2024)]%
        {sun2024umie}
\bibfield{author}{\bibinfo{person}{Lin Sun}, \bibinfo{person}{Kai Zhang}, \bibinfo{person}{Qingyuan Li}, {and} \bibinfo{person}{Renze Lou}.} \bibinfo{year}{2024}\natexlab{}.
\newblock \showarticletitle{Umie: Unified multimodal information extraction with instruction tuning}. In \bibinfo{booktitle}{\emph{Proceedings of the AAAI Conference on Artificial Intelligence}}, Vol.~\bibinfo{volume}{38}. \bibinfo{pages}{19062--19070}.
\newblock


\bibitem[Wang et~al\mbox{.}(2023)]%
        {10.1145/3581783.3612322}
\bibfield{author}{\bibinfo{person}{Jieming Wang}, \bibinfo{person}{Ziyan Li}, \bibinfo{person}{Jianfei Yu}, \bibinfo{person}{Li Yang}, {and} \bibinfo{person}{Rui Xia}.} \bibinfo{year}{2023}\natexlab{}.
\newblock \showarticletitle{Fine-Grained Multimodal Named Entity Recognition and Grounding with a Generative Framework}. In \bibinfo{booktitle}{\emph{Proceedings of the 31st ACM International Conference on Multimedia}} (Ottawa ON, Canada) \emph{(\bibinfo{series}{MM '23})}. \bibinfo{publisher}{Association for Computing Machinery}, \bibinfo{address}{New York, NY, USA}, \bibinfo{pages}{3934–3943}.
\newblock
\showISBNx{9798400701085}
\href{https://doi.org/10.1145/3581783.3612322}{doi:\nolinkurl{10.1145/3581783.3612322}}


\bibitem[Wu et~al\mbox{.}(2024)]%
        {wu_deepseek-vl2_2024}
\bibfield{author}{\bibinfo{person}{Zhiyu Wu}, \bibinfo{person}{Xiaokang Chen}, \bibinfo{person}{Zizheng Pan}, \bibinfo{person}{Xingchao Liu}, \bibinfo{person}{Wen Liu}, \bibinfo{person}{Damai Dai}, \bibinfo{person}{Huazuo Gao}, \bibinfo{person}{Yiyang Ma}, \bibinfo{person}{Chengyue Wu}, \bibinfo{person}{Bingxuan Wang}, \bibinfo{person}{Zhenda Xie}, \bibinfo{person}{Yu Wu}, \bibinfo{person}{Kai Hu}, \bibinfo{person}{Jiawei Wang}, \bibinfo{person}{Yaofeng Sun}, \bibinfo{person}{Yukun Li}, \bibinfo{person}{Yishi Piao}, \bibinfo{person}{Kang Guan}, \bibinfo{person}{Aixin Liu}, \bibinfo{person}{Xin Xie}, \bibinfo{person}{Yuxiang You}, \bibinfo{person}{Kai Dong}, \bibinfo{person}{Xingkai Yu}, \bibinfo{person}{Haowei Zhang}, \bibinfo{person}{Liang Zhao}, \bibinfo{person}{Yisong Wang}, {and} \bibinfo{person}{Chong Ruan}.} \bibinfo{year}{2024}\natexlab{}.
\newblock \bibinfo{title}{{DeepSeek}-{VL2}: {Mixture}-of-{Experts} {Vision}-{Language} {Models} for {Advanced} {Multimodal} {Understanding}}.
\newblock
\href{https://doi.org/10.48550/arXiv.2412.10302}{doi:\nolinkurl{10.48550/arXiv.2412.10302}}
\newblock
\shownote{arXiv:2412.10302 [cs]}.


\bibitem[Xiang and Wang(2019)]%
        {xiang2019survey}
\bibfield{author}{\bibinfo{person}{Wei Xiang} {and} \bibinfo{person}{Bang Wang}.} \bibinfo{year}{2019}\natexlab{}.
\newblock \showarticletitle{A survey of event extraction from text}.
\newblock \bibinfo{journal}{\emph{IEEE Access}}  \bibinfo{volume}{7} (\bibinfo{year}{2019}), \bibinfo{pages}{173111--173137}.
\newblock


\bibitem[Xu et~al\mbox{.}(2024)]%
        {xu2024large}
\bibfield{author}{\bibinfo{person}{Derong Xu}, \bibinfo{person}{Wei Chen}, \bibinfo{person}{Wenjun Peng}, \bibinfo{person}{Chao Zhang}, \bibinfo{person}{Tong Xu}, \bibinfo{person}{Xiangyu Zhao}, \bibinfo{person}{Xian Wu}, \bibinfo{person}{Yefeng Zheng}, \bibinfo{person}{Yang Wang}, {and} \bibinfo{person}{Enhong Chen}.} \bibinfo{year}{2024}\natexlab{}.
\newblock \showarticletitle{Large language models for generative information extraction: A survey}.
\newblock \bibinfo{journal}{\emph{Frontiers of Computer Science}} \bibinfo{volume}{18}, \bibinfo{number}{6} (\bibinfo{year}{2024}), \bibinfo{pages}{186357}.
\newblock


\bibitem[Yu et~al\mbox{.}(2020)]%
        {yu-etal-2020-improving-multimodal}
\bibfield{author}{\bibinfo{person}{Jianfei Yu}, \bibinfo{person}{Jing Jiang}, \bibinfo{person}{Li Yang}, {and} \bibinfo{person}{Rui Xia}.} \bibinfo{year}{2020}\natexlab{}.
\newblock \showarticletitle{Improving Multimodal Named Entity Recognition via Entity Span Detection with Unified Multimodal Transformer}. In \bibinfo{booktitle}{\emph{Proceedings of the 58th Annual Meeting of the Association for Computational Linguistics}}, \bibfield{editor}{\bibinfo{person}{Dan Jurafsky}, \bibinfo{person}{Joyce Chai}, \bibinfo{person}{Natalie Schluter}, {and} \bibinfo{person}{Joel Tetreault}} (Eds.). \bibinfo{publisher}{Association for Computational Linguistics}, \bibinfo{address}{Online}, \bibinfo{pages}{3342--3352}.
\newblock
\href{https://doi.org/10.18653/v1/2020.acl-main.306}{doi:\nolinkurl{10.18653/v1/2020.acl-main.306}}


\bibitem[Yu et~al\mbox{.}(2023)]%
        {yu-etal-2023-grounded}
\bibfield{author}{\bibinfo{person}{Jianfei Yu}, \bibinfo{person}{Ziyan Li}, \bibinfo{person}{Jieming Wang}, {and} \bibinfo{person}{Rui Xia}.} \bibinfo{year}{2023}\natexlab{}.
\newblock \showarticletitle{Grounded Multimodal Named Entity Recognition on Social Media}. In \bibinfo{booktitle}{\emph{Proceedings of the 61st Annual Meeting of the Association for Computational Linguistics (Volume 1: Long Papers)}}, \bibfield{editor}{\bibinfo{person}{Anna Rogers}, \bibinfo{person}{Jordan Boyd-Graber}, {and} \bibinfo{person}{Naoaki Okazaki}} (Eds.). \bibinfo{publisher}{Association for Computational Linguistics}, \bibinfo{address}{Toronto, Canada}, \bibinfo{pages}{9141--9154}.
\newblock
\href{https://doi.org/10.18653/v1/2023.acl-long.508}{doi:\nolinkurl{10.18653/v1/2023.acl-long.508}}


\bibitem[Zhang et~al\mbox{.}(2018)]%
        {zhang2018adaptive}
\bibfield{author}{\bibinfo{person}{Qi Zhang}, \bibinfo{person}{Jinlan Fu}, \bibinfo{person}{Xiaoyu Liu}, {and} \bibinfo{person}{Xuanjing Huang}.} \bibinfo{year}{2018}\natexlab{}.
\newblock \showarticletitle{Adaptive co-attention network for named entity recognition in tweets}. In \bibinfo{booktitle}{\emph{Proceedings of the AAAI conference on artificial intelligence}}, Vol.~\bibinfo{volume}{32}.
\newblock


\bibitem[Zhang et~al\mbox{.}(2017)]%
        {zhang2017improving}
\bibfield{author}{\bibinfo{person}{Tongtao Zhang}, \bibinfo{person}{Spencer Whitehead}, \bibinfo{person}{Hanwang Zhang}, \bibinfo{person}{Hongzhi Li}, \bibinfo{person}{Joseph Ellis}, \bibinfo{person}{Lifu Huang}, \bibinfo{person}{Wei Liu}, \bibinfo{person}{Heng Ji}, {and} \bibinfo{person}{Shih-Fu Chang}.} \bibinfo{year}{2017}\natexlab{}.
\newblock \showarticletitle{Improving event extraction via multimodal integration}. In \bibinfo{booktitle}{\emph{Proceedings of the 25th ACM international conference on Multimedia}}. \bibinfo{pages}{270--278}.
\newblock


\bibitem[Zhao et~al\mbox{.}(2024)]%
        {zhao2024comprehensive}
\bibfield{author}{\bibinfo{person}{Xiaoyan Zhao}, \bibinfo{person}{Yang Deng}, \bibinfo{person}{Min Yang}, \bibinfo{person}{Lingzhi Wang}, \bibinfo{person}{Rui Zhang}, \bibinfo{person}{Hong Cheng}, \bibinfo{person}{Wai Lam}, \bibinfo{person}{Ying Shen}, {and} \bibinfo{person}{Ruifeng Xu}.} \bibinfo{year}{2024}\natexlab{}.
\newblock \showarticletitle{A comprehensive survey on relation extraction: Recent advances and new frontiers}.
\newblock \bibinfo{journal}{\emph{Comput. Surveys}} \bibinfo{volume}{56}, \bibinfo{number}{11} (\bibinfo{year}{2024}), \bibinfo{pages}{1--39}.
\newblock


\bibitem[Zheng et~al\mbox{.}(2021)]%
        {zheng2021multimodal}
\bibfield{author}{\bibinfo{person}{Changmeng Zheng}, \bibinfo{person}{Junhao Feng}, \bibinfo{person}{Ze Fu}, \bibinfo{person}{Yi Cai}, \bibinfo{person}{Qing Li}, {and} \bibinfo{person}{Tao Wang}.} \bibinfo{year}{2021}\natexlab{}.
\newblock \showarticletitle{Multimodal relation extraction with efficient graph alignment}. In \bibinfo{booktitle}{\emph{Proceedings of the 29th ACM international conference on multimedia}}. \bibinfo{pages}{5298--5306}.
\newblock


\end{thebibliography}

\appendix



\end{document}